\pdfoutput=1

\documentclass[11pt]{article}

\usepackage[]{ACL2023}

\usepackage{times}
\usepackage{latexsym}
\usepackage{diagbox}
\usepackage{multirow}
\usepackage{float}
\usepackage{geometry}
\usepackage{tabularx}

\usepackage[T1]{fontenc}

\usepackage[utf8]{inputenc}

\usepackage{microtype}

\usepackage{inconsolata}
\usepackage{booktabs}
\usepackage{graphicx}
\usepackage{xurl}

\usepackage{caption}
\usepackage{subcaption}
\usepackage{calc}

\title{UKP-SQuARE v3: A Platform for Multi-Agent QA Research}

\author{
\begin{minipage}[t]{\textwidth}
\centering
\normalsize
\bf
Haritz Puerto,
Tim Baumgärtner,
Rachneet Sachdeva,
Haishuo Fang,
Hao Zhang,\\
Sewin Tariverdian,
Kexin Wang,
Iryna Gurevych \\
{\footnotesize \normalfont 
Ubiquitous Knowledge Processing Lab (UKP Lab), \\Department of Computer Science and Hessian Center for AI (hessian.AI), \\Technical University of Darmstadt \\
\url{www.ukp.tu-darmstadt.de}
} 
\end{minipage}
}

\begin{document}
\maketitle
\begin{abstract}
The continuous development of Question Answering (QA) datasets has drawn the research community's attention toward multi-domain models. A popular approach is to use \textit{multi-dataset models}, which are models trained on multiple datasets to learn their regularities and prevent overfitting to a single dataset. However, with the proliferation of QA models in online repositories such as GitHub or Hugging Face, an alternative is becoming viable. Recent works have demonstrated that combining expert agents can yield large performance gains over multi-dataset models. To ease research in \textit{multi-agent models}, we extend UKP-SQuARE, an online platform for QA research, to support three families of multi-agent systems: i) agent selection, ii) early-fusion of agents, and iii) late-fusion of agents. We conduct experiments to evaluate their inference speed and discuss the performance vs. speed trade-off compared to multi-dataset models. UKP-SQuARE is open-source\footnote{\href{https://github.com/UKP-SQuARE/square-core}{https://github.com/UKP-SQuARE/square-core}} and publicly available at \mbox{\href{https://square.ukp-lab.de}{square.ukp-lab.de}}.
\end{abstract}


\begin{figure*}
    \centering
    \includegraphics[width=\textwidth]{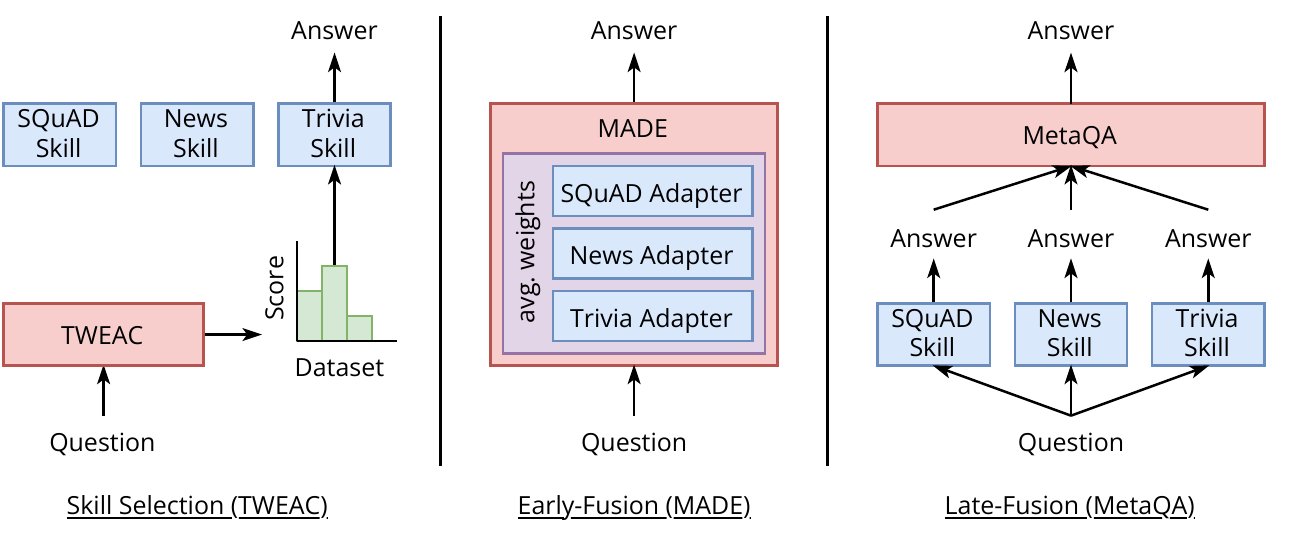}
    \caption{Overview of different multi-agent system architectures deployed in UKP-SQuARE. TWEAC (left) selects an agent (a \textit{Skill} in UKP-SQuARE) based on which dataset it predicts the question is closest to and on which dataset a Skill was trained. MADE (center) fuses the weights of adapters trained on different datasets. MetaQA (right) predicts the final answer from a set of answers and their confidence scores. We illustrate the architectures with three different Skills. However, in practice, more Skills are used. 
    }
    \label{fig:three-models}
\end{figure*}

\section{Introduction}

The current high-speed development of Artificial Intelligence yields thousands of datasets and trained models in repositories such as GitHub and Hugging Face \citep{qa_explosion}. These models are creating new research and application opportunities, such as high-performing Question Answering (QA) skills in chatbots \citep{burtsev-etal-2018-deeppavlov, miller-etal-2017-parlai}.
Comparing and analyzing these models usually requires learning libraries, writing code to run the models, and unifying their formats to compare them, which makes this process time-consuming and not scalable. 

UKP-SQuARE \citep{baumgartner-etal-2022-ukp, sachdeva-etal-2022-ukp} addresses this challenge, providing the first online platform that offers an ecosystem for QA research enabling reproducibility, analysis, and comparison of QA models through a standardized interface and from multiple angles (i.e., general behavior, explainability, adversarial attacks, and behavioral tests). 

The large variety of tasks and domains in QA datasets is pushing the research community towards creating models that generalize across domains \citep{fisch-etal-2019-mrqa, talmor-berant-2019-multiqa, khashabi-etal-2020-unifiedqa}. Currently, there are two main approaches to achieve this: i) multi-dataset models and ii) multi-agent models. While the former trains a model on multiple datasets \citep{talmor-berant-2019-multiqa, khashabi-etal-2020-unifiedqa}, the latter combines multiple expert agents \citep{geigle2021tweac, friedman-etal-2021-single, puerto-etal-2023-metaqa}. Concurrently, large language models (LLM) such as \mbox{GPT-3} \citep{brown2020language} are emerging as new powerful systems for multi-task and multi-domain NLP applications. These LLM models are complementary to the focus of our work, multi-agent systems. While LLMs show impressive performance, they are extremely expensive to run and can usually only be accessed through APIs or deployed with great hardware resources. On the other hand, multi-agent systems offer a solution to create multi-domain models reusing available pretrained models that can be run on more modest hardware, which is an important requirement, e.g. where data cannot be sent to third parties.

Multi-agent models are particularly promising due to the thousands of models readily available on online model hubs and their current exponential growth.\footnote{\href{https://www.nazneenrajani.com/emnlp_keynote.pdf}{https://www.nazneenrajani.com/emnlp\_keynote.pdf}} This growth in the number of models is increasing the interest of the community in multi-agent model research \citep{wang-etal-2020-multi-domain-named, Matena2021MergingMW, geigle2021tweac, friedman-etal-2021-single, puerto-etal-2023-metaqa, Wortsman2022ModelSA, jin2023dataless}. However, model hubs such as Hugging Face only allow inference on individual models, disregarding the possibility of combining them to make systems modular and multi-domain. This is a severe limitation as \citet{puerto-etal-2023-metaqa} showed that combining several QA models can yield performance gains of over 10 percentage points with respect to multi-dataset models (i.e., a single model trained on multiple datasets).


Therefore, we extend UKP-SQuARE to democratize access and research to multi-agent models. In particular, we add support to the three main methods to combine agents\footnote{An agent is referred to as \textit{Skill} in UKP-SQuARE.}: i) Skill selection, ii) early-fusion of Skills, and iii) late-fusion of Skills. The first consists of identifying the Skill with the highest likelihood of giving the correct answer and then routing the input to that Skill. We deploy TWEAC \citep[Transformer With Extendable QA Agent Classifiers;][]{geigle2021tweac} as an example of this method. The second one combines multiple models' weights to obtain a new model with the distributional knowledge of the source weights. We deploy MADE \citep[Multi-Adapter Dataset Experts;][]{friedman-etal-2021-single} as an example of this method. Lastly, the late-fusion of models consists of running multiple models to get their predictions and then combing them. This creates a system that can combine heterogeneous expert agents without reducing their performance in each domain. We provide MetaQA \citep{puerto-etal-2023-metaqa} as an example of this method.

UKP-SQuARE facilitates research on multi-agent QA systems by offering a platform equipped with dozens of agents and three methods to combine them. This upgrade holds paramount significance as the number of QA models created annually is increasing exponentially. UKP-SQuARE enables users to run, compare, and evaluate the strengths and weaknesses of multi-agent models, and compare them with multi-dataset models.

\section{Related Work}
The most famous types of multi-agent systems are Mixture of Experts (MoE) and ensemble methods. MoE consists of a gating mechanism that routes the input to a set of agents \citep{jacobs1991adaptive} while ensemble methods aggregate the outputs of multiple experts through a voting mechanism \citep{breiman1996bagging, freund1996experiments}. Much work has been made to simplify the training of these multi-agent systems \citep{scikit-learn, Chen_2016, he2021fastmoe, tutel}. However, as far as we know, there are no online platforms to run and compare them.

The most similar works to ours are the online model hubs such as Hugging Face's Model Hub\footnote{\href{https://huggingface.co/models}{https://huggingface.co/models}} and AdapterHub \citep{pfeiffer-etal-2020-adapterhub}. They both offer a large number of models to download. In addition, Hugging Face's Model Hub also allows running models through Spaces.\footnote{\href{https://huggingface.co/spaces}{https://huggingface.co/spaces}} However, this requires implementing the Space, which can be non-trivial for complex scenarios such as ours (i.e., deploying and comparing multi-agent systems). UKP-SQuARE removes technical barriers and allows researchers to deploy multi-agent systems with a user-friendly interface.

Transformer \citep{vaswani2017attention} models using adapters \citep{houlsby2019adapters} can also be seen as a type of multi-agent system. For this type of architecture, AdapterHub \citep{pfeiffer-etal-2020-adapterhub} is a well-established library. In addition to simplifying the training of adapter-based models, it allows composing adapters (i.e., agents) with methods such as AdapterFusion \citep{pfeiffer-etal-2021-adapterfusion} or stacking \citep{pfeiffer-etal-2020-mad}. However, this library is not an online platform for analyzing models such as UKP-SQuARE. Their focus is to offer tools to create models based on adapters.

\section{UKP-SQuARE}
UKP-SQuARE \citep{baumgartner-etal-2022-ukp, sachdeva-etal-2022-ukp} is the first online platform that offers an ecosystem for QA research. Its goal is to provide a common place to share, run, compare, and analyze QA models from multiple angles, such as explainability, adversarial attacks, behavioral tests, and I/O behaviors. The platform follows a flexible and scalable microservice architecture containing five main services:
\begin{itemize}
    \item \textbf{Datastores}: Provide access to collections of unstructured text such as Wikipedia and Knowledge Graphs such as ConceptNet \citep{speer-havasi-2012-representing}.
    \item \textbf{Models}: Enable the dynamic deployment and inference of any Transformer model that implements a Hugging Face pipeline \citep{wolf-etal-2020-transformers} including models that use the adapter-transformers \citep{pfeiffer-etal-2020-adapterhub} or sentence-transformers \citep{reimers-gurevych-2019-sentence} framework.
    \item \textbf{Skills}: central entity of the UKP-SQuARE. They specify a configurable QA pipeline (e.g., extractive, multiple-choice, and open-domain QA) leveraging Datastores and Models. Users interact with Skills since the platform's goal is to remove technical barriers and focus on QA research (i.e., the QA pipeline). These Skills are equivalent to agents in the multi-agent system literature.
    \item \textbf{Explainability}: Provides saliency maps, behavioral tests, and graph visualizations\footnote{For graph-based models.} that explains the outputs of a Skill.
    \item \textbf{Adversarial Attacks}: Create modified versions of the input to create adversarial attacks to expose vulnerabilities of the Skills.
\end{itemize}

All these services allow UKP-SQuARE to offer an ecosystem of tools to analyze Skills through a user-friendly interface without writing any code or complex configurations. UKP-SQuARE helps researchers identify the models' strengths and weaknesses to push the boundaries of QA research.

\subsection{Target Users and Scenarios}
This new update of UKP-SQuARE targets researchers working on multi-agent and multi-dataset systems. These users can use the platform as a showcase of their systems. The dozens of Skills already available in UKP-SQuARE simplify the deployment of multi-agent systems since users can employ our user-friendly interface to select the Skills they want to combine using the three families of methods we deploy. Furthermore, researchers can deploy their new multi-skill methods through a pull request in our repository.
The platform can also be used to analyze and compare multiple multi-agent systems from efficiency (i.e., inference time) and effectiveness (i.e., performance) points of view. Furthermore, it can also be used to compare multi-agent with multi-dataset systems. Lastly, UKP-SQuARE can also be used for teaching QA. The ecosystem of QA tools can be used to help students understand explainability, adversarial attacks, multi-dataset, and multi-agent models through interactive explanations with examples. Our platform can also be used to design homework where students train QA models and analyze them with the aforementioned QA tools.

\section{Multi-Agent Systems}
Multi-Agent systems are a type of multi-domain system that aggregate multiple expert agents from different domains to create a unified system. i.e., their focus is on the agents (\textit{Skills} in UKP-SQuARE). On the other hand, multi-dataset systems aim to learn a unified model from multiple data distributions to create a single, general agent. For example, UnifiedQA \citep{khashabi-etal-2020-unifiedqa} is a QA model trained on multiple datasets using a generative model to overcome format boundaries. 

However, \citet{2020t5} show that a model trained on multiple datasets may underperform the same architecture trained on a single dataset, i.e., multi-dataset models may underfit certain distributions. Based on this observation, \citet{puerto-etal-2023-metaqa} show that multi-agent models can avoid this limitation while being data-efficient to train and even outperform multi-dataset models by large margins in both in-domain and out-of-domain scenarios. This is possible because instead of using a very general architecture to solve multiple tasks, it uses a list of expert agents with specific architectures designed to solve those tasks (i.e., SOTA agents) and establishes a collaboration between these agents. However, this performance comes at a cost. The inference time is higher because it needs to run more than one model (at least one expert agent and one answer aggregator). 

Therefore, we extend UKP-SQuARE to add support to the three main approaches for multi-agent systems, which we refer to as Meta-Skills on the platform: i) Skill Selection (\S\ref{sec:tweac}), ii) Early-Fusion of Skills (\S\ref{sec:made}), and iii) Late-Fusion of Skills (\S\ref{sec:metaqa}). An overview of the different architectures is illustrated in Figure \ref{fig:three-models}.

\begin{figure}[t]
\centering
\includegraphics[width=\linewidth]{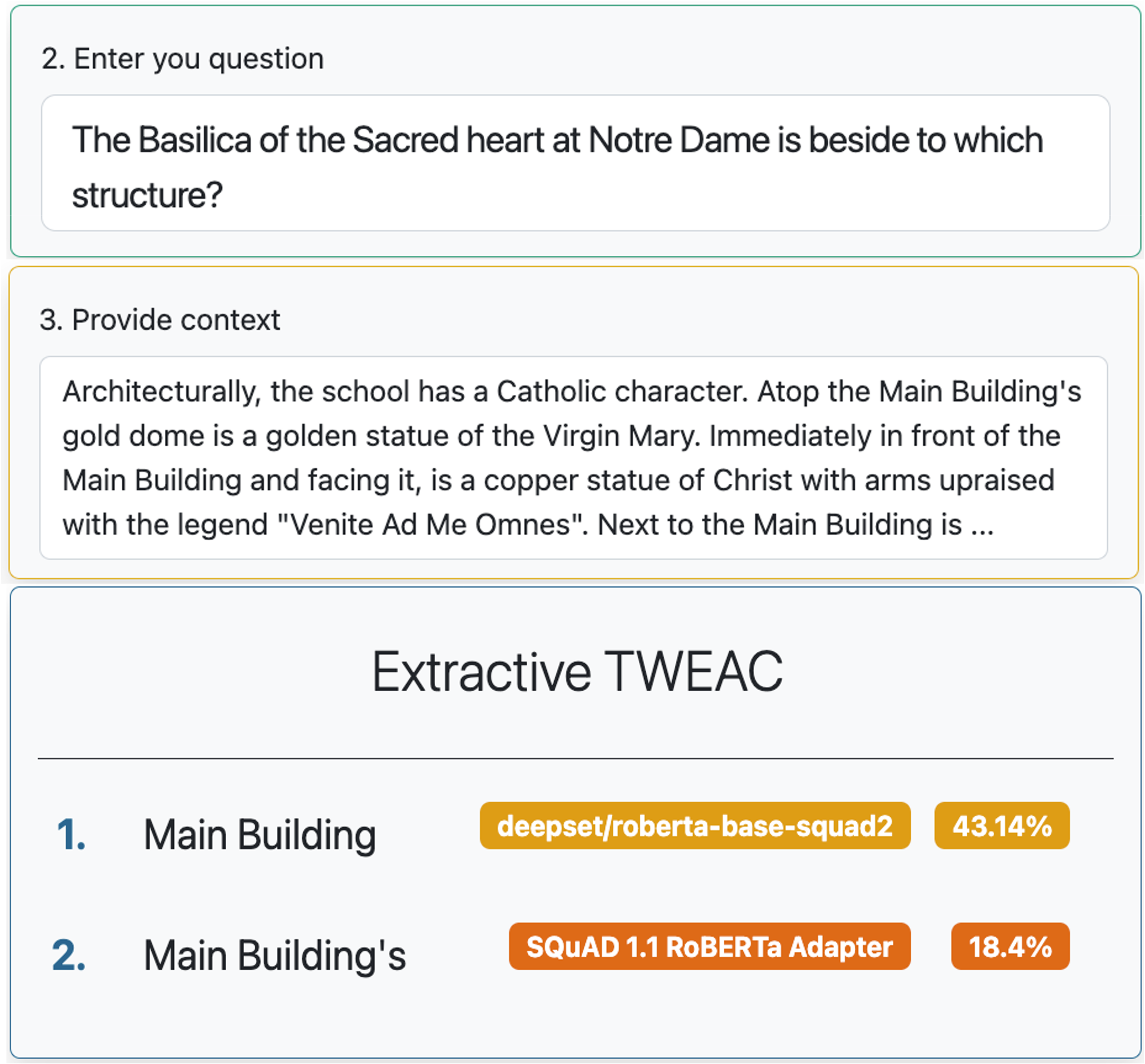}
\caption{TWEAC predicts that the question is \textit{SQuAD-like} and routes it to Skills trained on this dataset.}
\label{fig:tweac}
\end{figure}

\subsection{Skill Selection}\label{sec:tweac}
Skill selection is the simplest method of the three. It aims to identify the Skill with the highest likelihood of returning the correct answer to the input question and then route the input to that Skill. More formally, it defines a function $f: Q \to S$ that maps any question $Q$ to an available Skill $S$. \citet{geigle2021tweac} follow this approach and propose TWEAC (Transformer with Extendable QA Agent Classifiers), a Transformer model with a classification head for each Skill that maps questions to Skills. However, instead of predicting Skills, they predict \textit{datasets}, i.e., they identify the dataset from which the input question comes. Then, they select a Skill trained on that dataset. Using this method, they report a Skill prediction accuracy higher than 90\% across ten different QA types.

We train TWEAC on 16 datasets (shown in Appendix \ref{sec:appendix}) with an accuracy of 79\% and deploy it in UKP-SQuARE. The cause of the accuracy difference is the selection of the datasets. While the authors experiment on widely different QA tasks such as SQuAD, CommunityQA, and Weather Report, we use the most popular QA datasets, including the 2019 MRQA Shared Task \citep{fisch-etal-2019-mrqa}, which are more similar and thus, the task becomes more challenging since it is more difficult to distinguish the type of questions. We deploy two TWEAC Skills on UKP-SQuARE: one for extractive QA and another for multiple-choice. Figure \ref{fig:tweac} shows an extractive QA TWEAC that identifies the question as \textit{SQuAD-like} and routes it to two Skills trained on SQuAD.

\subsection{Early-Fusion of Skills}\label{sec:made}
This method combines the weights of multiple models to create a new model that generalizes across all the input models. 

\citet{friedman-etal-2021-single} propose to train adapter weights for individual datasets while sharing the weights of a common Transformer that is also trained with those adapters. Later, in a second training phase, they freeze the Transformer weights and fine-tune each adapter on its corresponding dataset. The intuition behind this is that the shared parameters encode the regularities of the QA task while the adapters model the sub-distributions. This training schema yields a model that performs robustly on new domains by averaging its adapter weights.

\begin{figure}[t]
\centering
\includegraphics[width=\linewidth]{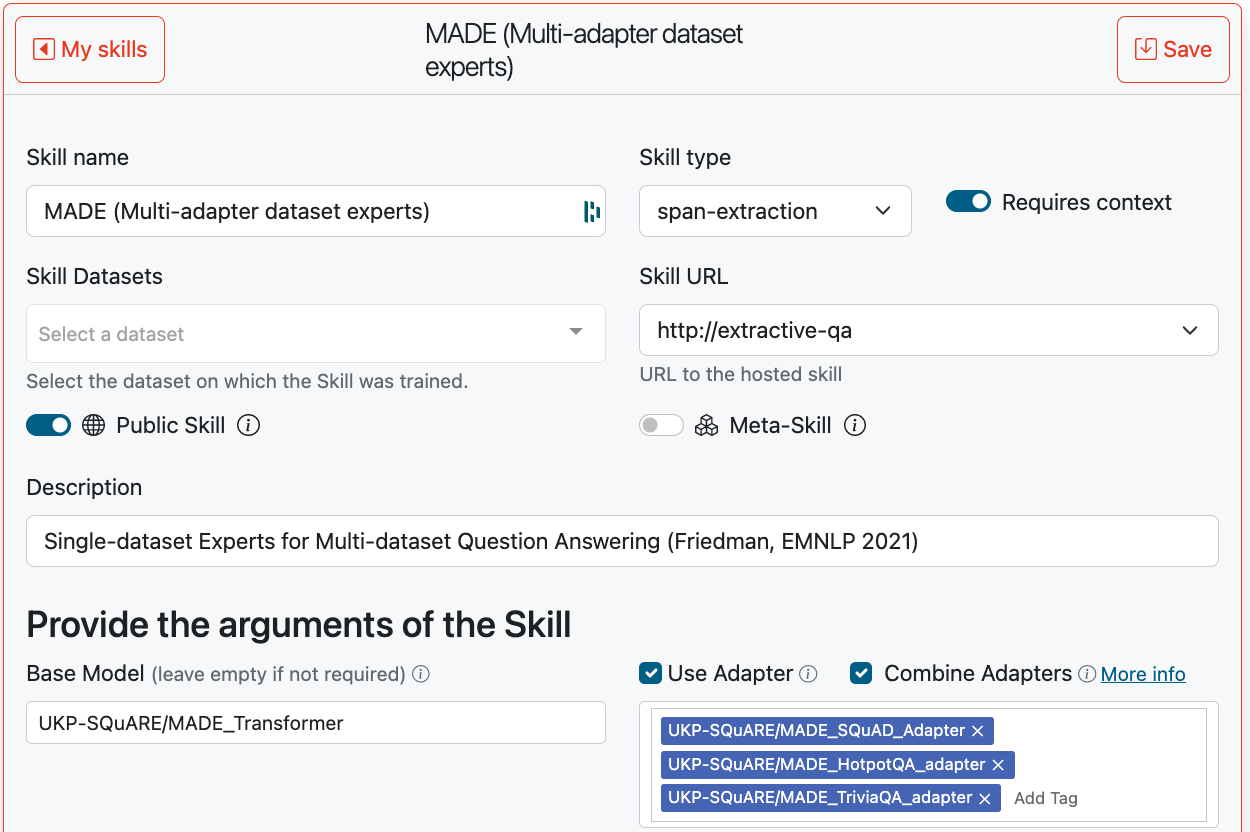}
\caption{UKP-SQuARE allows combining adapters by simply writing the list of adapters.}
\label{fig:skillcreation}
\end{figure}

Following this work, we extend UKP-SQuARE to allow the creation of Skills that average the weights of a series of adapters. To do this, on the Skill creation page (Figure \ref{fig:skillcreation}), users are prompted to select whether they wish to combine adapters and, if affirmative, which ones to average.

\subsection{Late-Fusion of Skills}\label{sec:metaqa}
\begin{figure}[t]
\centering
\includegraphics[width=\linewidth]{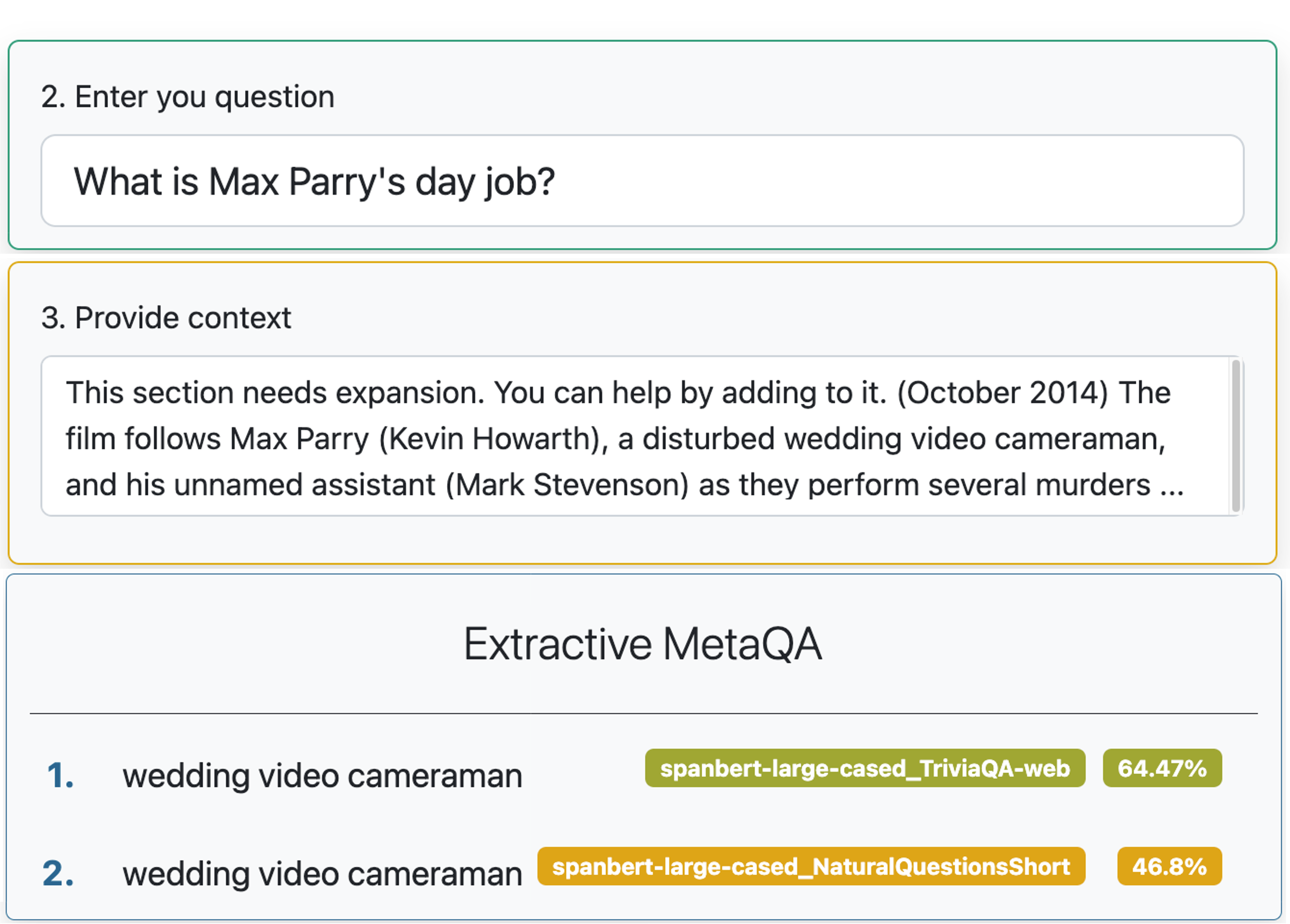}
\caption{UKP-SQuARE simplifies the analysis of the collaboration between the agents. The question comes from the DuoRC dataset. However, while the in-domain agent gives a wrong answer (not shown), MetaQA selects an out-of-domain agent that gives a correct answer. Only an excerpt of the context is shown.}
\label{fig:metaqa}
\end{figure}

Lastly, \citet{puerto-etal-2023-metaqa} propose MetaQA, a system that combines 18 heterogeneous expert agents across multiple formats. This system yields significant gains over multi-dataset models because some tasks require particular architectures to solve them, such as DROP \citep{dua-etal-2019-drop}, which requires numerical reasoning. Thus, while a \textit{one-size-fits-all} architecture cannot learn such a wide variety of distributions, a multi-agent system that combines predictions can use expert agents to solve these datasets and yield a higher-performing model in general. Figure \ref{fig:metaqa} shows how MetaQA answers a question from the \textit{DuoRC} dataset but selects an out-of-domain (OOD) agent instead of the in-domain agent to answer, which gives a wrong answer. Thanks to the interface provided by UKP-SQuARE, it is easier to analyze the collaboration between the Skills established by MetaQA.

One limitation of this type of system is its need to run multiple models, which makes it more expensive than the previous two approaches. To alleviate this limitation, we run the expert agents in parallel. In this way, the inference time of MetaQA remains close to the other multi-agent systems, as shown in Table~\ref{table:main_result}.

\subsection{Comparison of Multi-Skill Models}
\label{sec:comparison}
In this section, we compare the inference time of the deployed multi-skill systems (i.e., MetaQA, TWEAC, and MADE) and UnifiedQA as a representative of the multi-dataset models. We extract 20 random questions from the six datasets from the MRQA 2019 Shared Task \citep{fisch-etal-2019-mrqa} yielding a total of 120 questions and measure the time needed by each Skill to solve them. We repeat this process with five different random seeds and show the means and standard deviations in Table~\ref{table:main_result}. Each model has 8 CPUs\footnote{AMD EPYC 7543 with 2.8GHz.} assigned to it and runs behind an asynchronous API.

As shown in Table~\ref{table:main_result}, MetaQA is the slowest model. This is expected since it needs to run all the expert agents to get the predictions. However, its inference time is remarkably close to both MADE and TWEAC. TWEAC is surprisingly as fast as MADE, considering that TWEAC has to run at least two models (router and expert agent), while MADE only runs one. We conjecture that MADE is not faster because the adapter layers increase the depth of the transformer stack. UnifiedQA is the fastest model, as expected, since it is a multi-dataset model and hence, does not need to combine multiple agents.

Beyond inference, training time and cost are also interesting factors to consider. TWEAC and MetaQA are considered cheap to train assuming the existence of pretrained agents on online model hubs such as the Hugging Face Model Hub.\footnote{3.6K models on \url{https://huggingface.co/models?pipeline_tag=question-answering&sort=downloads}. Accessed on Feb 2023} Hence, the part that they train is a small router or answer aggregator. On the other hand, MADE and UnifiedQA require training a neural network from scratch in the task of question answering, which is much more challenging than simply routing questions or aggregating answers. Therefore, MADE and UnifiedQA need more training data than TWEAC and MetaQA, making them more expensive.

\begin{table}[t]
\centering
    

\begin{tabular}{lccc}
     \toprule
     \textbf{Model} & $\mathbf{F_1}$ & \textbf{\begin{tabular}[x]{@{}c@{}}Inference\\Time [s]\end{tabular}} & \textbf{Training} \\ \midrule
     TWEAC & 77.65 & 5.38 ± 0.06 & cheap \\
     MADE & 82.20 & 5.45 ± 0.18 & expensive \\
     MetaQA & 81.13 & 7.08 ± 0.16 & cheap \\
     UnifiedQA & 77.30 & 2.15 ± 0.02 & expensive\\
     \bottomrule
\end{tabular}

\caption{Comparison of inference time on UKP-SQuARE averaged over 600 predictions. Performance from their respective papers.
}
\label{table:main_result}
\end{table}

Table~\ref{table:main_result} shows the trade-off between performance, training, and inference efficiency. Although MetaQA is the slowest Skill to run, its inference time is very close to the other models' thanks to the parallel inference of the expert agents offered by UKP-SQuARE (cf. Figure~\ref{fig:three-models}). Furthermore, it is cheap to train, has almost the highest performance, and is compatible with any QA format. This makes it interesting for scenarios where model updating, performance, and flexibility are vital. TWEAC is also cheap and as flexible as MetaQA. Although, it is significantly worse than MetaQA on extractive QA datasets. This makes TWEAC ideal in the same scenarios as MetaQA but where running the expert agents in parallel is difficult (i.e., when MetaQA cannot be used). MADE has the highest performance and is as fast as TWEAC. However, it is more expensive to train than MetaQA and TWEAC, and it is not as flexible as MetaQA and TWEAC since it cannot be used for multiple formats simultaneously. Therefore, it should be used when inference, performance, and simple deployment are vital, while the model is not expected to need re-training (i.e., updates) often and is not required to be compatible with multiple QA formats at the same time. Lastly, UnifiedQA is compatible with any text-based QA format but has lower (although competitive) results. Although it is the fastest to run, it is more expensive to train than TWEAC and MetaQA. Thus, its ideal use case is a scenario where a simple deployment is needed while being flexible to process any text-based QA format. Therefore, this small study suggests that in scenarios where new domains are introduced often, router-based systems such as MetaQA might be more suitable, whereas, in scenarios where inference speed or simple deployment are needed, MADE and UnifiedQA might be more appropriate.



\section{Conclusions and Discussions}
In this work, we have extended UKP-SQuARE to support multi-agent models. In particular, we deployed a routing system, TWEAC \citep{geigle2021tweac}, a method to combine adapter weights, MADE \citep{friedman-etal-2021-single}, and a model that combines the prediction of multiple Skills, MetaQA \citep{puerto-etal-2023-metaqa}. We have conducted experiments on these three models and UnifiedQA \citep{khashabi-etal-2020-unifiedqa}, a multi-dataset system, to analyze the trade-off between the performance, efficiency, and flexibility of these systems. We showed that in scenarios where new domains or expertise are often needed, MetaQA provides the best trade-off since its performance is close to the best model, it is compatible with any QA format, cheap to train, and its inference runtime is close to TWEAC and MADE using the parallel engine provided by UKP-SQuARE. However, when simple deployment is needed or the model is not expected to be updated, MADE and UnifiedQA might be more appropriate.

This update of UKP-SQuARE is of utmost importance due to the current speed of development of QA models that creates thousands of models per year. Our platform eases the deployment, running, comparison, and analysis of QA Skills. With this update, we also facilitated the aggregation of these Skills into Multi-Skills simplifying research on multi-agent systems. We leave as future work the comparison of these modular systems with prompting-based QA in large language models \citep{brown2020language, zhong-etal-2022-proqa}.

\section*{Limitations}
UKP-SQuARE v3 does not aim to provide all existing multi-skill systems off the shelf. Instead, we deploy three different approaches and encourage the community to share, deploy and compare their multi-skill systems. Using the modular Skill system of UKP-SQuARE and the reference implementations, users can reconfigure the existing multi-skill pipelines or implement and deploy their own through a streamlined pull request.\footnote{For details, we refer to the documentation at \href{https://square.ukp-lab.de/docs/home/components/skills}{https://square.ukp-lab.de/docs/home/components/skills}}

Another limitation is that the multi-skill systems deployed in this paper have been shown to work effectively with no more than ~20 Skills. Hence, the effectiveness of multi-skill systems remains unknown for a larger number of Skills. We hope that UKP-SQuARE v3 can help shed light on this topic.

Lastly, since multi-skill systems combine several models, it is feasible that the resulting system can inherit biases and unfair behaviors. Although the Skills we used are not intended to exhibit any bias or unfairness, users should use them at their own discretion.

\section*{Ethics Statement}

\paragraph{Intended Use}
The intended use of UKP-SQuARE v3 is deploying, running, comparing, analyzing, and combining Skills. Our platform provides dozens of Skills readily available to be combined using the implemented multi-agent systems or new systems to be created by the community. This simplifies the analysis of these systems and thus fosters multi-agent QA research.

\paragraph{Potential Misuse}
A malicious user could train multiple Skills with biased and unfair behaviors, such as a QA system that responds harmful answers, and combine them with the deployed methods available in UKP-SQuARE. UKP-SQuARE does not provide any Skill with such an intended behavior, but the community is free to upload any model to our platform. Therefore, we encourage the community not to publicly upload such models unless there is a clear research intention with a discussion of the ethics of such research, and in this case, make the Skills private, so that nobody can use them in an unintended way. We are not liable for errors, false, biased, offensive, or any other unintended behavior of the Skills. Users should use them at their own discretion.

\paragraph{Environmental Impact}
The use of UKP-SQuARE can reduce the computational cost of reproducing prior research since it prevents the community from training models that are already trained. 

\section*{Acknowledgements}
We thank Haris Jabbar, Martin Tutek, and Imbesat Hassan Rizvi for their insightful comments on a previous draft of this paper.

This work has been funded by the German Research Foundation (DFG) as part of the UKP-SQuARE project (grant GU 798/29-1), the QASciInf project (GU 798/18-3), and by the German Federal Ministry of Education and Research and the Hessian Ministry of Higher Education, Research, Science and the Arts (HMWK) within their joint support of the National Research Center for Applied Cybersecurity ATHENE.

\bibliography{anthology,custom}
\bibliographystyle{acl_natbib}
\clearpage
\appendix\label{sec:appendix}

\section{Further Updates}
Although our focus is the development of multi-domain systems, we further extended UKP-SQuARE with other minor but important features.

\subsection{Knowledge Graph Question Answering}
To maximize the advantage of multi-agent systems, UKP-SQuARE needs to be compatible with most QA systems available. UKP-SQuARE is already compatible with most QA formats (i.e., extractive, multiple-choice, boolean, and abstractive), and in this version, we add support for Knowledge Graph Question Answering (KGQA) systems. \citet{sachdeva-etal-2022-ukp} include support for neuro-symbolic systems that combine language models with ConceptNet but lack support for KGQA models. Thus, this paper implements a generic KGQA Skill compatible with any knowledge graph and any generation model that generates SPARQL queries. As a demo, we deploy a BART-based \citep{lewis-etal-2020-bart} semantic parser on KQA Pro \citep{cao-etal-2022-kqa}, which is a complex KGQA dataset based on Wikidata \citep{wikidata} with nine different question types. The KGQA Skill parses a question into an executable SPARQL query, which is then executed against a KG to get the final answer. For this purpose, a dataset-centric subgraph is deployed using virtuoso.\footnote{\url{https://virtuoso.openlinksw.com/}} Thanks to the modularity of UKP-SQuARE, we can flexibly combine different semantic parsers with different KGs to get the final answer. Therefore, all aforementioned multi-Skill methods can be easily adapted for multi-Skill KGQA, which we leave as future work.

\subsection{BERTViz}
\begin{figure}[h]
\begin{center}
\includegraphics[width=\linewidth]{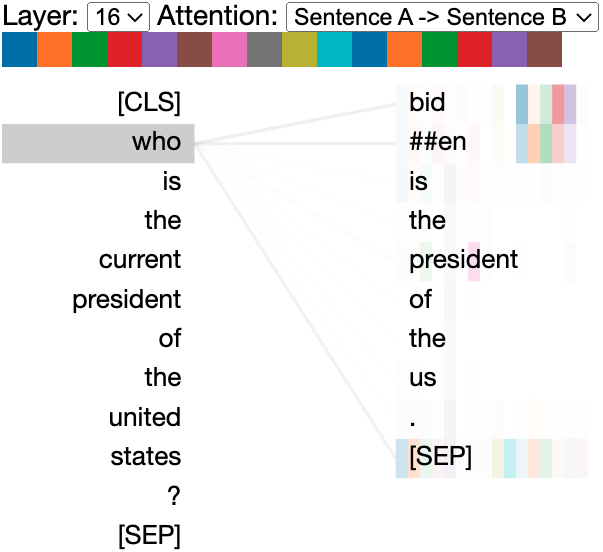}
\caption{Screenshot of the BERTViz attention visualization in UKP-SQuARE. The token \textit{who} attends to \textit{Biden} (the answer) more than to other tokens.}
\label{fig:bertviz}    
\end{center}
\end{figure}

We also extended our explainability ecosystem by adding BERTViz \citep{vig-2019-multiscale}, a method that allows the exploration of the attention weights as shown in Figure \ref{fig:bertviz}. While UKP-SQuARE v2 focuses on high-level explanations through saliency maps, BERTViz offers a low-level explanation utilizing the attention weights across all layers of the transformer models.

\subsection{Datastores}
Lastly, regarding Datastores, while UKP-SQuARE v1 focuses on document collections such as Wikipedia or PubMed, and UKP-SQuARE v2 focuses on Knowledge Graphs, UKP-SQuARE v3 offers a datastore that is updated in real-time. This type of information is vital for real-time questions (i.e., questions whose answers may change over time; \citet{kasai2022realtime}). For instance, some facts, such as the president of a country, can change quickly. Therefore, we deploy a real-time datastore by using the Bing Search API. This datastore does not store documents and, instead, relies on the Bing Search engine to retrieve online documents (i.e., websites) that are more likely to be updated. 

Furthermore, we create an information-retrieval Skill that allows the inference of only an IR model (instead of combining them with a reader model). We allow providing relevance-feedback for sparse retrieval, as it has shown to perform well in information-seeking and interactive scenarios \citep{baumgartner-etal-2022-incorporating}.

\onecolumn
\section{Datasets}
\setlength\extrarowheight{3pt}
\footnotesize



\begin{tabularx}{\textwidth}{llcccc}
\toprule
    
    \textbf{Dataset} & \textbf{Characteristics} & \textbf{Train} & \textbf{Dev} & \textbf{Test} & \textbf{License} \\
\midrule
      SQuAD \citep{rajpurkar-etal-2016-squad}        & Crowdsourced questions on Wikipedia  & 6573  & 5253 & 5254     & MIT \\
  NewsQA \citep{trischler-etal-2017-newsqa}       & Crowdsourced questions about News    & 74160  & 2106         & 2106 & MIT \\
  HotpotQA \citep{yang-etal-2018-hotpotqa}     & Crowdsourced multi-hop questions on Wikipedia & 72928 & 2950 & 2951 & MIT \\
  SearchQA \citep{dunn2017searchqa}     & Web Snippets, Trivia questions from J! Archive & 117384 & 8490 & 8490 & MIT \\
  NQ \citep{kwiatkowski-etal-2019-natural} & Wikipedia, real user queries on Google Search & 104071 & 6418 & 6418 & MIT \\
  TriviaQA-web \citep{joshi-etal-2017-triviaqa} & Web Snippets, crowdsourced trivia questions & 61688 & 3892 & 3893 & MIT \\
 

\bottomrule
\end{tabularx}
\captionof{table}{Summary of the datasets used in \S \ref{sec:comparison}.}
\label{table:datasets}


\section{Expert Agents}
\setlength\extrarowheight{3pt}
\footnotesize
\begin{tabularx}{\textwidth}{lp{0.8\linewidth}}
\toprule
\textbf{Expert Agents} &\textbf{Link}\\
\midrule
Span-BERT Large for SQuAD  & \href{https://huggingface.co/haritzpuerto/spanbert-large-cased_SQuAD}{https://huggingface.co/haritzpuerto/spanbert-large-cased\_SQuAD}       \\

Span-BERT Large for NewsQA & \href{https://huggingface.co/haritzpuerto/spanbert-large-cased_NewsQA}{https://huggingface.co/haritzpuerto/spanbert-large-cased\_NewsQA}         \\

Span-BERT Large for HotpotQA & \href{https://huggingface.co/haritzpuerto/spanbert-large-cased_HotpotQA}{https://huggingface.co/haritzpuerto/spanbert-large-cased\_HotpotQA} \\

Span-BERT Large for SearchQA & \href{https://huggingface.co/haritzpuerto/spanbert-large-cased_SearchQA}{https://huggingface.co/haritzpuerto/spanbert-large-cased\_SearchQA}  \\

Span-BERT Large for NQ & \href{https://huggingface.co/haritzpuerto/spanbert-large-cased_NaturalQuestionsShort}{https://huggingface.co/haritzpuerto/spanbert-large-cased\_NaturalQuestionsShort}   \\

Span-BERT Large for TriviaQA-web & \href{https://huggingface.co/haritzpuerto/spanbert-large-cased_TriviaQA-web}{https://huggingface.co/haritzpuerto/spanbert-large-cased\_TriviaQA-web}  \\

Span-BERT Large for QAMR & \href{https://huggingface.co/haritzpuerto/spanbert-large-cased_QAMR}{https://huggingface.co/haritzpuerto/spanbert-large-cased\_QAMR}   \\

Span-BERT Large for DuoRC & \href{https://huggingface.co/haritzpuerto/spanbert-large-cased_DuoRC}{https://huggingface.co/haritzpuerto/spanbert-large-cased\_DuoRC}  \\
\bottomrule
\end{tabularx}

\caption{List of the expert agents used for TWEAC and MetaQA.}
\label{table:agents}

\twocolumn

\end{document}